\documentclass{article}
\usepackage{spconf,amsmath,graphicx,booktabs,multirow,hyperref}
\hypersetup{hidelinks}
\newcommand{\dur}{\mathit{duration}}
\newcommand{\tstart}{\mathit{start}}
\newcommand{\tend}{\mathit{end}}
\newcommand{\secs}{\,\mathrm{s}}

\title{IndicFDB: Benchmarking Full-Duplex Voice Agents across Indian Languages}
\name{\parbox{6.8in}{\centering Rajarshi Roy$^{1}$, Shobhit Banga$^{2}$, Jonathan Raiman$^{1}$, Supriya Paul$^{2}$, Bhaskar Singh$^{2}$,\\
Manmeet Kaur$^{2}$, Sagar Jain$^{2}$, Hanuman Sidh$^{2}$, Pranav Sharma$^{2}$, Aditya Singh$^{2}$,\\
Aaditya Pareek$^{2}$, Manas Dhir$^{2}$, Adi Margolin$^{1}$, Niket Agarwal$^{1}$, Bryan Catanzaro$^{1}$}}
\address{$^{1}$NVIDIA \qquad $^{2}$VoiceArena}

\begin{document}
\maketitle
\begin{abstract}
\small
Full-duplex voice agents must handle pauses, take turns, backchannel, and respond to user interruptions in real time.
Full-Duplex-Bench evaluates these behaviors, but its English-only corpus and reliance on word-timestamped ASR and an English-prompted LLM judge make it difficult to extend to Indian languages.
We introduce IndicFDB, which extends it to ten languages spoken in India with 12,350 samples, nearly 17 times as many as the original.
We address three challenges: finding conversational events in multilingual speech, evaluating their timing without reliable word-level alignment, and judging responses across languages.
We mine pause handling, turn taking, and backchanneling samples from roughly 50,000 hours of channel-separated conversations using voice activity detection (VAD), and construct human-validated synthetic user interruption samples.
Language-independent VAD heuristics evaluate timing, while an open-weight transcription and translation pipeline converts responses to English for LLM ratings of relevance and quality.
Across seven voice agents, commercial APIs show unexpectedly consistent behavior across languages but are either fast or robust to pauses, never both, while monolingual open full-duplex models expose further tradeoffs among backchanneling, response quality, and latency.
\end{abstract}
\normalsize

\section{Introduction}
India, the world's most populous country~\cite{unpopulation}, is a major deployment setting for multilingual voice agents.
Agents there must converse across languages with distinct scripts, pronunciations, and conversational conventions.
Evaluating their real-time behavior is difficult because the automatic speech recognition (ASR), forced alignment, and semantic judging components that benchmarks rely on are unevenly supported across these languages.
A benchmark built on English word timestamps or an English-only judge cannot simply be translated and scaled.

Full-Duplex-Bench (FDB)~\cite{lin2025fdb} evaluates four categories of conversational dynamics: pause handling, turn taking, backchanneling, and user interruption.
FDB uses ASR with word-level timestamps to mine samples from human dialogue and to evaluate agent timing, and prompts GPT-4o in English to judge the relevance and quality of agent responses to user interruptions.
Ohashi et al.~\cite{ohashi2026multifaceted} instead use voice activity detection (VAD) to mine samples and define heuristic rewards that improve the same dynamics.

We introduce IndicFDB, a full-duplex benchmark covering ten languages spoken in India: Indian English, Hindi, Bengali, Punjabi, Gujarati, Marathi, Kannada, Telugu, Tamil, and Malayalam.
We retain FDB's four categories but build a far larger set of 12,350 samples (Table~\ref{tab:data}), using multilingual VAD~\cite{silero} for both mining and evaluation.
We also introduce a \textit{transcribe-translate-judge} pipeline built entirely from open-weight ASR~\cite{bodhan} and language~\cite{gemma4} models, which rates user interruption responses in all ten languages.
Evaluating four commercial APIs and three open models, we find that each commercial API behaves consistently across languages, yet no system is both fast and robust to pauses.
The benchmark samples and associated mining and evaluation code are intended for release at \url{https://3bk47sbkf7b.github.io/indicfdb/}.

\begin{table}[h]
\centering
\caption{Samples by conversational dynamics category.}
\label{tab:data}
%\footnotesize
\small
\setlength{\tabcolsep}{2pt}
\begin{tabular}{@{}lccccc@{}}
\toprule
Benchmark & \shortstack{Pause\\handling} & \shortstack{Turn\\taking} & \shortstack{Back\\channeling} & \shortstack{User\\interruption} & Total \\
\midrule
\shortstack[l]{Full-Duplex-\\Bench~\cite{lin2025fdb}} & 353 & 119 & 55 & 200 & 727 \\
\midrule
\shortstack[l]{IndicFDB\\(per language)} & 315 & 235 & 385 & 300 & 1,235 \\
\midrule
\shortstack[l]{IndicFDB\\(10 languages)} & 3,150 & 2,350 & 3,850 & 3,000 & 12,350 \\
\bottomrule
\end{tabular}
\end{table}

\section{Benchmark construction}
\subsection{Natural data curation}
\label{sec:natural_data_curation}
We automatically mine pause handling, backchanneling, and turn taking samples from VoiceArena's corpus of channel-separated, two-speaker conversations, which contains roughly 5,000 hours of speech in each of the ten languages.
Because FDB's word-level timestamping tools do not cover all ten languages~\cite{lin2025fdb}, we rely on VAD-based heuristics as in~\cite{ohashi2026multifaceted}.
Based on human validation feedback, we (i) refine the heuristics and (ii) add multi-threshold VAD filtering for high confidence.

\textbf{Speech categorization:} Silero VAD~\cite{silero} segments each speaker's channel into speech intervals called interpausal units (IPUs).
We consider one speaker at a time, labeling that speaker X and the other speaker Y.
Consecutive IPUs form utterances, split at silences longer than 0.5\,s, and consecutive utterances form turns, split at silences longer than 1.5\,s.
Silences of 0.5--1.5\,s are \textit{short pauses}, and longer silences are \textit{long pauses}.
The mining heuristics for each category are as follows:

\textbf{Pause handling:} We select a single turn of speaker X, $T_X$, that satisfies three conditions: (i) $\dur(T_X) \geq 8\secs$; (ii) it contains at least one \textit{short pause}; and (iii) speaker Y produces no speech.

\textbf{Backchanneling:} We select a single turn of speaker X, $T_X$, that satisfies two conditions: (i) $\dur(T_X) \geq 15\secs$; and (ii) it contains at least one short overlapping utterance (backchannel) $U_{Y_k}$ from speaker Y with $\dur(U_{Y_k}) \leq 1.2\secs$.

\textbf{Turn taking:} We select turns $T_X$ and $T_Y$ from speakers X and Y that satisfy three conditions: (i) $\dur(T_X) \geq 5\secs$ and $\dur(T_Y) \geq 5\secs$; (ii) $0\secs \leq \tstart(T_Y)-\tend(T_X) \leq 0.5\secs$; and (iii) no speech from Y during $T_X$ or from X during $T_Y$.

Mined clips are filtered to avoid overlapping examples from the same source conversation and are then reviewed by language experts for validity.

\textbf{Multi-threshold VAD filtering:} At the default VAD threshold of 0.5, non-speech sounds could be detected as speech or real speech may go undetected; lowering the threshold worsens the former, and raising it worsens the latter.
Running two VAD passes with low (0.1) and high (0.7) thresholds on a 1,000-hour multilingual corpus subset, we find 285,383 IPUs detected by both thresholds and an additional 25,971 IPUs (9.1\%) detected only by the low threshold.
The commonly detected IPUs are validated as speech, while the low-threshold-only IPUs include non-speech sounds and noisy, low-volume, or paralinguistic speech.
We therefore mine only segments in which neither speaker has low-threshold-only IPUs, so that every detected IPU is speech and every silence is free of low-confidence speech.

\subsection{Synthetic multilingual user interruption generation}
We first expand FDB's 200 English user interruption samples to 300.
Each sample pairs a user \textit{context} turn with an \textit{interrupt} turn that cuts into the agent's response to the context and should make the agent respond to the interruption instead.
Language experts translate the \textit{context} and \textit{interrupt} turns into the ten languages with the aid of machine translation.
Speech is synthesized with the Gemini 3.1 Flash TTS Preview API~\cite{google2026gemini31flashtts}.
Language experts validate every utterance, and the turns are stitched into a timed user interruption scenario following FDB.

\section{Multilingual evaluation}
As in FDB, we stream the user-side audio (speaker X) of each sample to each voice agent and record the agent's output for the duration of the input.
For mined samples, the other channel (speaker Y) serves as the ground-truth \textit{Human} reference.

\subsection{Conversational dynamics metrics}
As in Section~\ref{sec:natural_data_curation}, we use VAD to segment agent speech into utterances and turns, labeling utterances shorter than 1.2\,s as backchannels and utterances longer than 2\,s as takeovers.
FDB reports takeover rates, which are desirable in some categories and undesirable in others; we instead report a success rate for every category, so higher is always better.
The metrics for each category are as follows:

\textbf{Pause handling:} A sample succeeds if the agent never takes over, i.e., it waits through the user's short pauses.

\textbf{Backchanneling:} A sample succeeds if there are no takeovers.
\textit{Frequency} is the number of backchannels divided by the sample duration.
\textit{Jensen--Shannon distance (JSD)} compares agent backchannel timestamps with the ground-truth \textit{Human} backchannel timestamps, using the same 0.2\,s binning and smoothing as FDB.
JSD is 1 when the agent produces no backchannels.

\textbf{Turn taking:} After the user's turn $T_U$ ends at $\tend(T_U)$, any agent turn $T_A$ with $\dur(T_A) > 1\secs$ and $\tstart(T_A) > \tend(T_U)$ is labeled a response.
A sample succeeds if a response exists and no takeover occurs before $\tend(T_U)$.
Latency is $\tstart(T_A) - \tend(T_U)$ for the earliest response $T_A$.

\textbf{User interruption:} Let $T_U$ be the user's interrupt turn.
Agent IPUs overlapping $\tstart(T_U)$ answer the \textit{context} and are discarded; then any agent turn $T_A$ with $\dur(T_A) > 1\secs$ and $\tstart(T_A) > \tstart(T_U)$ is labeled a response.
A sample succeeds if a response exists.
Latency is $\max(0, \tstart(T_A) - \tend(T_U))$.

Across samples in a category, success rate (S.R.) is the mean success value, and latency is the mean over successful samples; backchannel frequency and JSD are averaged over all samples, independent of success.

\subsection{Response content evaluation}
For successful user interruption samples, we transcribe the agent's response audio with Bodhan Indic Transcribe Core~\cite{bodhan}, which supports all ten languages.
Gemma 4 31B IT~\cite{gemma4} translates non-English transcripts into English, while English transcripts pass through unchanged.
Gemma 4 31B IT then rates each English response against English versions of the context and interrupt turns on FDB's 0--5 relevance scale.
Judging in a single language keeps the judge prompt and rating scale identical across all ten languages.
We chose these components after qualitatively inspecting errors from alternatives, including Gemma 4 E4B IT~\cite{gemma4e4b} for transcription and IndicTrans2~\cite{gala2023indictrans} and Bodhan Indic Translate~\cite{indic-translate-2026} for translation.

\section{Results and discussion}
We evaluate four commercial voice APIs (GPT Live~\cite{openai2026gptlive1}, Gemini 3.8 Live~\cite{google2026gemini38live}, Gemini 3.1 Flash Live Preview~\cite{google2026gemini31flashlive}, and Grok Voice Think Fast 2.0~\cite{xai2026grokvoicethinkfast20}) and three open full-duplex models (PersonaPlex~\cite{personaplex}, Moshi~\cite{moshi}, and JoshTalks Human-1~\cite{human1-2026}).
Language coverage is unequal: GPT Live and Gemini 3.8 have outputs in ten languages, Gemini 3.1 in nine, Grok in three, PersonaPlex and Moshi in English, and Human-1 in Hindi.
Table~\ref{tab:language_results} therefore reports each available model-language pair rather than cross-language averages.

The commercial APIs occupy opposite ends of a latency--robustness tradeoff.
GPT Live responds after 0.069--0.282\,s on average, faster than the human reference in nine of ten languages, but often takes over when the user pauses: its success rate is 23.81--74.29\% for pause handling and 15.32--43.90\% for backchanneling.
Gemini 3.8, Gemini 3.1, and Grok succeed on at least 99.05\% of pause handling samples in every tested language, but respond after 1.125--1.400\,s, 1.747--2.197\,s, and 2.459--2.561\,s, respectively.
The same split holds for user interruptions, which GPT Live answers after 0.135--0.322\,s and the other APIs after 1.098--1.946\,s.
No evaluated system escapes this tradeoff: across all 35 model-language pairs, every pair with sub-second turn taking latency has at most 90.16\% pause handling success, and every pair above 99\% responds after at least 1.125\,s.
TurnBench reports a similar speed--selectivity tradeoff for standalone turn-taking detectors~\cite{jiang2026turnbench}.

Backchanneling success alone is insufficient, since silence never causes a takeover.
Gemini 3.8 succeeds on 98.18--100\% of backchanneling samples but produces only 0.009--0.027 backchannels/s against a human rate of 0.070--0.098/s, and Gemini 3.1 and Grok only 0.001--0.005/s; their JSD of 0.882--0.992 mostly reflects missing backchannels, which receive a JSD of 1.
GPT Live matches or exceeds the human rate in every language while taking over in most of the same samples, and some of its sub-1.2\,s utterances may be brief answers rather than backchannels.
PersonaPlex comes closest to avoiding takeovers at a human-like rate, with 63.12\% success at 0.091/s.

Although most of these languages have fewer public speech resources than English, each commercial API keeps the same behavioral profile across them.
Gemini 3.8 succeeds on every pause handling sample in all ten languages with turn taking latency within 1.125--1.400\,s, and the latency ordering GPT Live $<$ Gemini 3.8 $<$ Gemini 3.1 $<$ Grok holds wherever these systems are compared.
GPT Live's success rates vary more, e.g., from 23.81\% (Bengali) to 74.29\% (Tamil) for pause handling, but mined samples and human behavior also differ across languages (human backchannels: 0.070--0.098/s), so natural-sample metrics are best compared within a language.
The parallel user interruption samples permit a more controlled comparison: Grok, Gemini 3.8, and Gemini 3.1 ratings stay within 4.787--4.997, whereas GPT Live falls from 4.980 in English to 4.550 in Punjabi and 4.003 in Gujarati.
As Gemini 3.8 and 3.1 Gujarati responses pass through the same pipeline and receive 4.953 and 4.900, this gap more likely reflects GPT Live's responses than the pipeline.

The open models show that timing and content metrics can rank systems differently.
PersonaPlex is the only pair combining sub-second turn taking latency (0.360\,s) with over 75\% pause handling success (90.16\%), and it has the lowest English JSD (0.799) with a 4.843 rating.
Moshi takes over in 96.88\% of backchanneling samples and has the lowest English rating (3.791).
In Hindi, Human-1 exceeds GPT Live in pause handling (72.06 versus 61.59\%) and backchanneling success (45.97 versus 15.32\%) and has the lowest JSD (0.795), yet responds after 2.021\,s with a 0.564 rating; since the same pipeline rates the other Hindi systems at 4.857--4.973, transcription and translation errors are unlikely to explain this gap fully.

\section{Conclusion and future work}
IndicFDB makes full-duplex evaluation practical across ten languages spoken in India.
Its 12,350 samples, nearly 17 times as many as FDB, are mined from roughly 50,000 hours of conversation with alignment-free, multi-threshold VAD heuristics and validated by language experts, and are scored by language-independent timing heuristics and a transcribe-translate-judge pipeline built entirely from open-weight models.
Across seven voice agents, commercial APIs behave consistently across languages but favor either speed or pause robustness, and the best backchannel timing comes from open models that trail on robustness, latency, or response quality.
No system yet combines human-like latency, pause robustness, natural backchanneling, and high-quality responses.

Two limitations remain: VAD heuristics may mislabel corner cases, such as brief answers as backchannels, and transcription and translation errors can accumulate before judging.
Beyond evaluation, the corpus and mining heuristics could supply multilingual training data and rewards for interactivity alignment~\cite{ohashi2026multifaceted}, and the mined samples could benchmark standalone turn-taking models in these languages, as TurnBench~\cite{jiang2026turnbench} does for end-of-turn and interruption detection.
Releasing the samples and scripts will support these directions and extension to further languages.

\clearpage
% Generated from evaluation_summary.json by generate_results_table.py.
\begin{table*}[p]
\centering
\caption{Per-language evaluation. S.R.: success rate (\%); Lat.: latency (s); JSD: Jensen--Shannon distance; Freq.: backchannels/s; Rating: user interruption response rating (0--5). Human references have no user interruption output.}
\label{tab:language_results}
\normalsize
\setlength{\tabcolsep}{6pt}
\renewcommand{\arraystretch}{0.90}
\begin{tabular}{@{}llrrrrrrrrr@{}}
\toprule
 & & \multicolumn{1}{c}{Pause handling} & \multicolumn{2}{c}{Turn taking} & \multicolumn{3}{c}{Backchanneling} & \multicolumn{3}{c}{User interruption} \\
\cmidrule(lr){3-3}\cmidrule(lr){4-5}\cmidrule(lr){6-8}\cmidrule(lr){9-11}
Language & Model & S.R.$\uparrow$ & S.R.$\uparrow$ & Lat.$\downarrow$ & S.R.$\uparrow$ & JSD$\downarrow$ & Freq.$\uparrow$ & S.R.$\uparrow$ & Lat.$\downarrow$ & Rating$\uparrow$ \\
\midrule
\multirow{7}{*}{\shortstack[l]{\textbf{English}\\\textbf{(Indian)}}} & Human & 100.00 & 100.00 & 0.273 & 100.00 & 0.000 & 0.070 & -- & -- & -- \\
 & GPT Live & 64.13 & 93.19 & \textbf{0.282} & 42.86 & 0.826 & \textbf{0.076} & 99.67 & \textbf{0.135} & 4.980 \\
 & Grok & 99.05 & 92.34 & 2.459 & 98.70 & 0.989 & 0.002 & \textbf{100.00} & 1.739 & 4.923 \\
 & Gemini 3.8 & \textbf{100.00} & \textbf{100.00} & 1.338 & \textbf{99.48} & 0.953 & 0.009 & \textbf{100.00} & 1.157 & 4.973 \\
 & Gemini 3.1 & \textbf{100.00} & 99.15 & 1.747 & 98.96 & 0.981 & 0.003 & \textbf{100.00} & 1.485 & \textbf{4.997} \\
 & PersonaPlex & 90.16 & \textbf{100.00} & 0.360 & 63.12 & \textbf{0.799} & 0.091 & \textbf{100.00} & 0.307 & 4.843 \\
 & Moshi & 57.46 & 71.06 & 1.475 & 3.12 & 0.870 & 0.056 & 99.00 & 0.449 & 3.791 \\
\midrule
\multirow{6}{*}{\textbf{Hindi}} & Human & 100.00 & 100.00 & 0.259 & 100.00 & 0.000 & 0.079 & -- & -- & -- \\
 & GPT Live & 61.59 & 91.06 & \textbf{0.181} & 15.32 & 0.822 & \textbf{0.109} & \textbf{100.00} & \textbf{0.322} & 4.857 \\
 & Grok & 99.05 & 85.11 & 2.556 & 98.18 & 0.986 & 0.004 & 99.67 & 1.628 & 4.960 \\
 & Gemini 3.8 & \textbf{100.00} & \textbf{99.15} & 1.400 & 98.18 & 0.942 & 0.010 & \textbf{100.00} & 1.255 & 4.970 \\
 & Gemini 3.1 & \textbf{100.00} & 98.30 & 2.128 & \textbf{100.00} & 0.990 & 0.003 & \textbf{100.00} & 1.678 & \textbf{4.973} \\
 & Human-1 & 72.06 & 89.36 & 2.021 & 45.97 & \textbf{0.795} & 0.110 & 95.67 & 1.573 & 0.564 \\
\midrule
\multirow{5}{*}{\textbf{Bengali}} & Human & 100.00 & 100.00 & 0.306 & 100.00 & 0.000 & 0.079 & -- & -- & -- \\
 & GPT Live & 23.81 & \textbf{100.00} & \textbf{0.189} & 37.92 & \textbf{0.801} & \textbf{0.115} & \textbf{100.00} & \textbf{0.216} & 4.853 \\
 & Grok & 99.68 & 91.49 & 2.561 & 98.44 & 0.992 & 0.002 & 99.67 & 1.946 & 4.860 \\
 & Gemini 3.8 & \textbf{100.00} & \textbf{100.00} & 1.282 & \textbf{100.00} & 0.946 & 0.011 & \textbf{100.00} & 1.140 & 4.903 \\
 & Gemini 3.1 & 99.68 & 99.57 & 2.045 & \textbf{100.00} & 0.991 & 0.001 & 99.33 & 1.574 & \textbf{4.960} \\
\midrule
\multirow{3}{*}{\textbf{Punjabi}} & Human & 100.00 & 100.00 & 0.245 & 100.00 & 0.000 & 0.095 & -- & -- & -- \\
 & GPT Live & 56.19 & \textbf{100.00} & \textbf{0.137} & 43.90 & \textbf{0.804} & \textbf{0.123} & \textbf{100.00} & \textbf{0.308} & 4.550 \\
 & Gemini 3.8 & \textbf{100.00} & \textbf{100.00} & 1.142 & \textbf{100.00} & 0.921 & 0.015 & \textbf{100.00} & 1.193 & \textbf{4.907} \\
\midrule
\multirow{4}{*}{\textbf{Gujarati}} & Human & 100.00 & 100.00 & 0.289 & 100.00 & 0.000 & 0.089 & -- & -- & -- \\
 & GPT Live & 73.97 & \textbf{100.00} & \textbf{0.069} & 35.84 & \textbf{0.814} & \textbf{0.106} & \textbf{100.00} & \textbf{0.247} & 4.003 \\
 & Gemini 3.8 & \textbf{100.00} & \textbf{100.00} & 1.129 & 99.48 & 0.931 & 0.014 & \textbf{100.00} & 1.217 & \textbf{4.953} \\
 & Gemini 3.1 & 99.68 & 99.15 & 2.057 & \textbf{100.00} & 0.987 & 0.002 & \textbf{100.00} & 1.728 & 4.900 \\
\midrule
\multirow{4}{*}{\textbf{Marathi}} & Human & 100.00 & 100.00 & 0.284 & 100.00 & 0.000 & 0.082 & -- & -- & -- \\
 & GPT Live & 54.29 & \textbf{100.00} & \textbf{0.107} & 37.14 & \textbf{0.808} & \textbf{0.128} & \textbf{100.00} & \textbf{0.260} & 4.850 \\
 & Gemini 3.8 & \textbf{100.00} & \textbf{100.00} & 1.161 & \textbf{100.00} & 0.947 & 0.012 & \textbf{100.00} & 1.158 & \textbf{4.973} \\
 & Gemini 3.1 & \textbf{100.00} & \textbf{100.00} & 1.883 & 99.22 & 0.981 & 0.003 & \textbf{100.00} & 1.563 & \textbf{4.973} \\
\midrule
\multirow{4}{*}{\textbf{Kannada}} & Human & 100.00 & 100.00 & 0.321 & 100.00 & 0.000 & 0.098 & -- & -- & -- \\
 & GPT Live & 37.14 & 88.09 & \textbf{0.174} & 15.58 & \textbf{0.830} & \textbf{0.097} & \textbf{100.00} & \textbf{0.175} & 4.680 \\
 & Gemini 3.8 & \textbf{100.00} & \textbf{100.00} & 1.185 & \textbf{100.00} & 0.882 & 0.027 & \textbf{100.00} & 1.141 & 4.920 \\
 & Gemini 3.1 & \textbf{100.00} & 99.57 & 2.197 & \textbf{100.00} & 0.987 & 0.003 & \textbf{100.00} & 1.625 & \textbf{4.923} \\
\midrule
\multirow{4}{*}{\textbf{Telugu}} & Human & 100.00 & 100.00 & 0.301 & 100.00 & 0.000 & 0.086 & -- & -- & -- \\
 & GPT Live & 61.90 & \textbf{99.57} & \textbf{0.090} & 41.56 & \textbf{0.797} & \textbf{0.116} & \textbf{100.00} & \textbf{0.168} & 4.837 \\
 & Gemini 3.8 & \textbf{100.00} & 99.15 & 1.190 & \textbf{99.74} & 0.898 & 0.019 & \textbf{100.00} & 1.098 & 4.877 \\
 & Gemini 3.1 & \textbf{100.00} & 98.72 & 1.884 & \textbf{99.74} & 0.976 & 0.005 & \textbf{100.00} & 1.578 & \textbf{4.980} \\
\midrule
\multirow{4}{*}{\textbf{Tamil}} & Human & 100.00 & 100.00 & 0.311 & 100.00 & 0.000 & 0.081 & -- & -- & -- \\
 & GPT Live & 74.29 & \textbf{100.00} & \textbf{0.173} & 15.58 & \textbf{0.817} & \textbf{0.110} & \textbf{100.00} & \textbf{0.189} & 4.763 \\
 & Gemini 3.8 & \textbf{100.00} & \textbf{100.00} & 1.147 & \textbf{100.00} & 0.941 & 0.015 & 99.67 & 1.113 & 4.906 \\
 & Gemini 3.1 & \textbf{100.00} & 98.72 & 1.934 & 99.22 & 0.980 & 0.005 & \textbf{100.00} & 1.494 & \textbf{4.947} \\
\midrule
\multirow{4}{*}{\textbf{Malayalam}} & Human & 100.00 & 100.00 & 0.289 & 100.00 & 0.000 & 0.087 & -- & -- & -- \\
 & GPT Live & 69.21 & \textbf{100.00} & \textbf{0.096} & 29.87 & \textbf{0.799} & \textbf{0.121} & \textbf{100.00} & \textbf{0.265} & 4.763 \\
 & Gemini 3.8 & \textbf{100.00} & \textbf{100.00} & 1.125 & \textbf{99.74} & 0.918 & 0.016 & \textbf{100.00} & 1.285 & 4.787 \\
 & Gemini 3.1 & 99.68 & 99.15 & 2.025 & \textbf{99.74} & 0.985 & 0.003 & \textbf{100.00} & 1.525 & \textbf{4.883} \\
\bottomrule
\end{tabular}
\end{table*}

\clearpage
\bibliographystyle{IEEEbib}
\bibliography{refs}
\end{document}